\documentclass[letterpaper, 10 pt, conference]{ieeeconf}  

\IEEEoverridecommandlockouts                              

\usepackage{amsmath}
\usepackage{amssymb}
\usepackage{graphicx} %
\usepackage{hyperref}
\usepackage{multirow} 
\usepackage{subcaption} 
\usepackage{silence} 
\usepackage{cite}
\usepackage{tabularx}
\usepackage{makecell}

\usepackage{tcolorbox}
\newtcbox{\revibox}{colback = violet, coltext=white, boxrule = 0pt,arc=5pt, boxsep=0pt,left=2pt,right=2pt,top=2pt, bottom=2pt, on line}
\usepackage[colorinlistoftodos]{todonotes}
\newtcbox{\reviiibox}{colback = orange, coltext=white, boxrule = 0pt,arc=5pt, boxsep=0pt,left=3pt,right=3pt,top=2pt, bottom=2pt, on line}

\title{\LARGE \bf
StereoMamba: Real-time and Robust Intraoperative Stereo Disparity Estimation via Long-range Spatial Dependencies 
\author{Xu Wang, Jialang Xu, Shuai Zhang, Baoru Huang, Danail Stoyanov, and Evangelos B. Mazomenos
\thanks{This work was supported in whole, or in part, by the EPSRC-funded UCL Centre for Doctoral Training in Intelligent, Integrated Imaging in Healthcare (i4health) [EP/S021930/1]; the EPSRC Human-centric Machine Intelligence to optimise Robotic Surgical Training "HuMIRoS" Project [EP/Z534754/1]; the UCL Research Excellence Scholarships programme; the NIHR UCLH Biomedical Research Centre [NIHR203328]; the Department of Science, Innovation and Technology (DSIT) and the Royal Academy of Engineering under the Chair in Emerging Technologies programme. For the purpose of open access, the author has applied a CC BY public copyright licence to any author accepted manuscript version arising from this submission. \textit{(Corresponding authors: X. Wang and E. B. Mazomenos.)}}
\thanks{All authors are with the UCL Hawkes Institute, London, UK. X. Wang, J. Xu, and E. B. Mazomenos are with the Department of Medical Physics and Biomedical Engineering, University College London, London, UK. S. Zhang, B. Huang and D. Stoyanov are with the Department of Computer Science, University College London, London, UK.
{\tt\small \{xu.wang.23; jialang.xu.22; shuai.z; baoru.huang; danail.stoyanov; e.mazomenos\}@ucl.ac.uk}}%
}
}

\begin{document}

\bstctlcite{IEEEexample:BSTcontrol}

\maketitle
\thispagestyle{empty}
\pagestyle{empty}

\begin{abstract}
Stereo disparity estimation is crucial for obtaining depth information in robot-assisted minimally invasive surgery (RAMIS). While current deep learning methods have made significant advancements, challenges remain in achieving an optimal balance between accuracy, robustness, and inference speed. To address these challenges, we propose the StereoMamba architecture, which is specifically designed for stereo disparity estimation in RAMIS. Our approach is based on a novel Feature Extraction Mamba (FE-Mamba) module, which enhances long-range spatial dependencies both within and across stereo images. To effectively integrate multi-scale features from FE-Mamba, we then introduce a novel Multidimensional Feature Fusion (MFF) module. Experiments against the state-of-the-art on the \textit{ex-vivo} SCARED benchmark demonstrate that StereoMamba achieves superior performance on EPE of 2.64 px and depth MAE of 2.55 mm, the second-best performance on Bad2 of 41.49\% and Bad3 of 26.99\%, while maintaining an inference speed of 21.28 FPS for a pair of high-resolution images (1280$\times$1024), striking the optimum balance between accuracy, robustness, and efficiency. Furthermore, by comparing synthesized right images, generated from warping left images using the generated disparity maps, with the actual right image, StereoMamba achieves the best average SSIM (0.8970) and PSNR (16.0761), exhibiting strong zero-shot generalization on the \textit{in-vivo} RIS2017 and StereoMIS datasets. Code is available at: \href{https://github.com/MichaelWangGo/StereoMamba}{StereoMamba}
\end{abstract}


\section{INTRODUCTION}
Stereo endoscopes are routinely employed in robotic-assisted minimally invasive surgery (RAMIS) to visualize the internal anatomy, providing surgeons with depth perception for precise instrument manipulation\cite{psychogyiosMSDESISMultitaskStereo2022}. Accurate, real-time, and robust disparity estimation from stereo video is a critical component in RAMIS, for understanding the geometry of the surgical scene and enabling downstream tasks such as preoperative image registration\cite{CHEN2023107121} and intraoperative navigation\cite{FENG2024103026}. However, ensuring these capabilities especially in \textit{in-vivo} environments, presents many challenges\cite{chenSpatiotemporalLayersBased2024}.

\begin{figure}[!t]
    \centering
    \includegraphics[width=\linewidth]{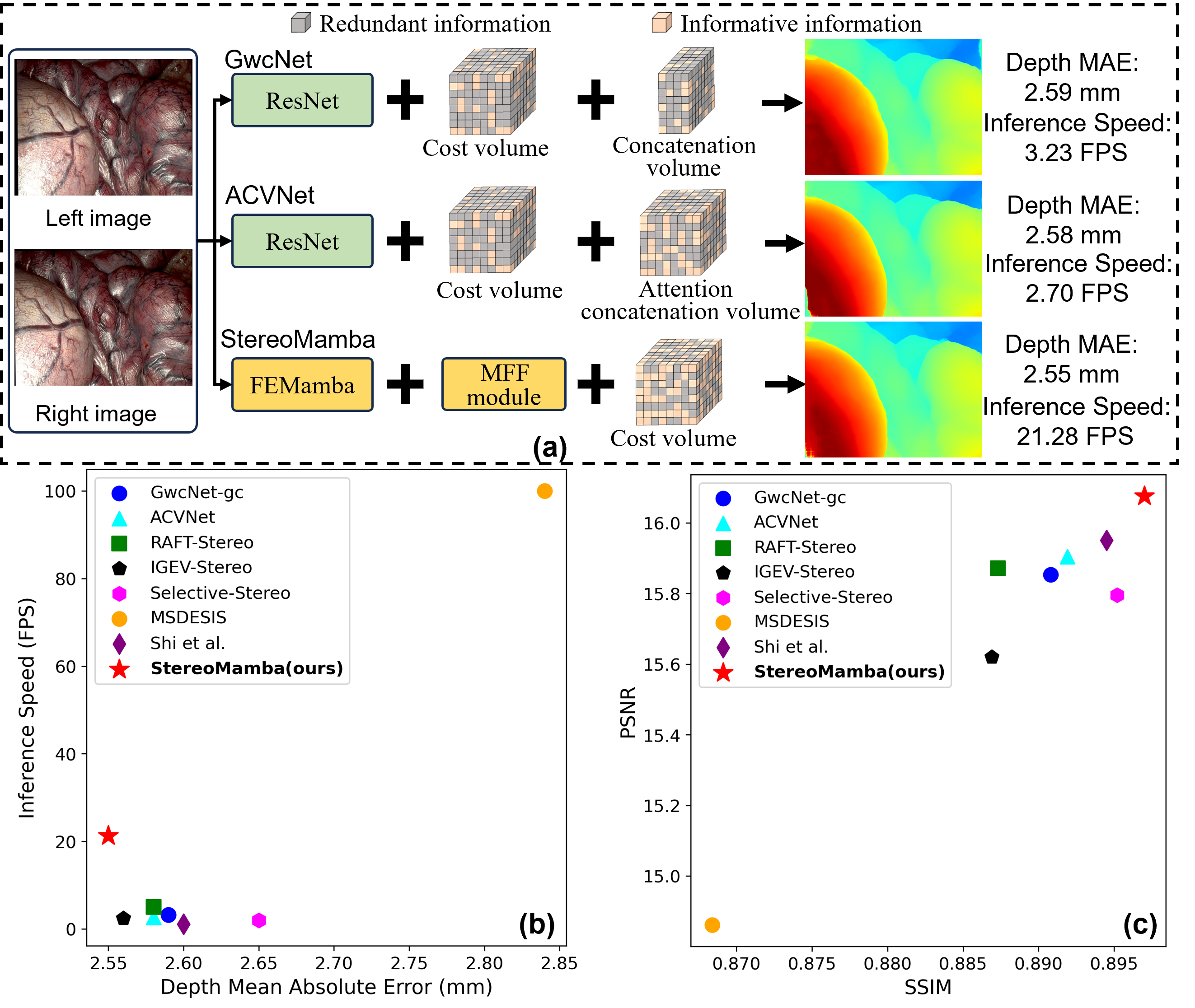}
    \caption{Comparisons with SOTA disparity estimation methods. (a) shows the different methodologies between GwcNet\cite{guoGroupwiseCorrelationStereo2019}, ACVNet\cite{xuAttentionConcatenationVolume2022}, and our proposed StereoMamba. (b) illustrates the trade-off between accuracy and inference speed. (c) compares the similarity between synthesized right images (generated from left images and disparity maps on two unseen \textit{in-vivo} datasets) and actual right images.}
    \label{fig1}
\end{figure}

Stereo disparity estimation is fundamentally a matching task, where each pixel in the rectified left image searches along the epipolar line in the rectified right image to find the optimal corresponding pixel with the lowest matching cost. A typical stereo matching pipeline\cite{cheng2020hierarchical} can be split into three or four steps: feature extraction, cost volume construction, disparity estimation and/or refinement. For cost volume construction, many state-of-the-art (SOTA) approaches\cite{guoGroupwiseCorrelationStereo2019,xuAttentionConcatenationVolume2022,wang2024selective} first fix a maximum disparity range (typically 192 px is applied), and then compute the matching cost between the left and right feature maps for each disparity value starting from 0, creating a 4D cost volume (height $\times$ width $\times$ disparity $\times$ feature dimension). Then, the disparity value that minimizes the matching cost for each pixel pair is found within the 4D cost volume. However, relying solely on pixel-level features often leads to ambiguity, especially in RAMIS scenes that contain large textureless regions or repetitive patterns, where multiple pixel matches may appear equally optimal. To alleviate this issue, it is crucial to incorporate both fine-grained \textit{local} features—captured from shallow Convolutional Neural Networks (CNNs)—and \textit{global} features—representing broader spatial features across the image captured from deeper CNN layers—during the feature extraction stage\cite{changPyramidStereoMatching2018}.

PSMNet enhances stereo matching by leveraging global context information through spatial pyramid pooling (SPP)\cite{changPyramidStereoMatching2018} and dilated convolutions, which extend pixel-level features to region-level representations across multiple scales. Another line of work focuses on improving cost volume construction effectiveness to further improve stereo matching accuracy. For example, GwcNet\cite{guoGroupwiseCorrelationStereo2019} introduces a novel group-wise correlation mechanism to refine the cost volume, providing more precise matching features for disparity regression, shown in Fig.~\ref{fig1}(a). 
However, these methods are limited by the receptive field of CNNs, which only extract features within single images. For stereo matching, establishing cross-image connections is essential for accurate correspondence retrieval\cite{dingTransMVSNetGlobalContextaware2022}. ACVNet\cite{xuAttentionConcatenationVolume2022} also adopts CNNs for feature extraction, but introduces an attention concatenation volume that links left and right features during cost volume construction, which generates attention weights based on relevant cues in order to suppress redundant information and to enhance informative information in the cost volume, shown in Fig.~\ref{fig1}(a). We propose that introducing these cross-image connections earlier—at the feature extraction stage—can improve the performance of stereo matching (shown in Fig.~\ref{fig1}).

Transformer architectures could be a potential solution with their capacity to model long-range spatial dependencies\cite{dosovitskiyImageWorth16x162021}. The self-attention mechanism enables the extraction of global contextual features within a single image, while cross-attention facilitates correspondence between stereo image pairs. 
Cheng et al.\cite{cheng2022deep} investigate Transformer integration within stereo matching pipelines\cite{cheng2020hierarchical}, demonstrating that leveraging Transformers for feature representation learning and CNNs for cost aggregation leads to faster convergence, improved accuracy, and enhanced generalization. However, Transformer-based stereo matching methods often suffer from quadratic computational and memory complexity due to the Query-Key product\cite{dosovitskiyImageWorth16x162021}. This poses a significant limitation in real-time \textit{in-vivo} applications, where both high accuracy and efficiency are critical. Striking a balance between estimation accuracy and model complexity remains a key challenge in advancing stereo disparity estimation.

Mamba\cite{guMambaLinearTimeSequence}, a selective State Space Model (SSM)\cite{guEFFICIENTLYMODELINGLONG2022a} architecture originally proposed for natural language modelling, which combines the strengths of CNN's linear complexity and transformers' long-range spatial dependencies, is a promising alternative for sequence modelling. Inspired by these advancements, several SSM-based visual backbone networks, such as VMamba\cite{liuVMambaVisualState2024} and Vision Mamba\cite{zhuVisionMambaEfficient2024}, have been proposed for image classification\cite{10604894} and segmentation\cite{MA2024112203}. However, no work has explored Mamba's unique capabilities for accurate, fast, and robust stereo disparity estimation in RAMIS, which remains an ongoing challenge in the field.

To this end, we propose StereoMamba, a novel end-to-end deep neural network designed for stereo disparity estimation in RAMIS. It features a powerful Feature Extraction Mamba (FE-Mamba) module that leverages both self-attention and cross-attention to enhance long-range spatial dependencies within and cross stereo images. To effectively integrate multi-scale features from FE-Mamba to the stereo matching pipeline, we introduce a Multidimensional Feature Fusion (MFF) module, which seamlessly combines self-attentive and cross-attentive features. The fused features are then divided into multiple groups along the channel dimension, where each left feature group is cross-correlated with its corresponding right feature group across all disparity levels, generating group-wise correlation maps. These maps are subsequently aggregated to construct the final cost volume, which is processed by a disparity regression network to generate the final disparity maps.

Our main contributions are the following:
\begin{itemize}
    
    \item \textbf{A novel feature extraction and fusion mechanism for stereo disparity estimation:} We propose FE-Mamba for feature extraction and MFF for multi-scale feature integration (Fig.~\ref{fig1}(a)). Compared to existing SOTA methods GwcNet\cite{guoGroupwiseCorrelationStereo2019} and ACVNet\cite{xuAttentionConcatenationVolume2022}, StereoMamba’s feature extractor and fusion strategy provide more informative features for cost volume construction.

    \item \textbf{A balance between accuracy, robustness, and inference speed:} StereoMamba achieves a state-of-the-art EPE of 2.64 px and depth MAE of 2.55 mm on the SCARED benchmark while maintaining real-time inference at 21.28 FPS (Fig.~\ref{fig1}(b)).

    \item \textbf{Strong zero-shot generalization on unseen surgical datasets }: By synthesizing right images using estimated disparity maps and original left images, StereoMamba achieves superior SSIM score of 0.8970, demonstrating high generalization ability (Fig.~\ref{fig1}(c)).
    
\end{itemize}

\section{RELATED WORK}

Iterative methods such as Recurrent All-Pairs Field Transforms (RAFT),  build a 4D cost volume by computing correlations between all pixel pairs, and iteratively updating the flow field using a Gated Recurrent Unit (GRU)-based update operator\cite{teedRAFTRecurrentAllPairs2020}. RAFT-Stereo\cite{lipsonRAFTStereoMultilevelRecurrent2021} extends this approach by employing multi-level Convolutional GRU (ConvGRU) to iteratively update the disparity field using local cost values retrieved from All-Pairs Correlations (APC). However, APCs lack global information and struggle with local ambiguities in challenging regions. In contrast, IGEV-Stereo introduces a novel module that encodes non-local geometry, contextual information, and local matching details, enhancing the effectiveness of each ConvGRU iteration\cite{xuIterativeGeometryEncoding2023}. This provides a better initial disparity map to the ConvGRUs, resulting in faster convergence. Selective-Stereo\cite{wang2024selective} proposes Selective Recurrent Unit (SRU) module that can adaptively fuse hidden disparity information at multiple frequencies for edge and smooth regions, and derive the Contextual Spatial Attention (CSA) module to help information from different receptive fields and frequencies fuse, resulting in outperforming RAFT-Stereo and IGEV-Stereo. However these modules increase the sizes of convolutional kernels, leading to high memory and time costs.

MSDESIS\cite{psychogyiosMSDESISMultitaskStereo2022} proposes a multi-task network for surgical instrument segmentation and stereo disparity estimation, demonstrating that supervising the segmentation task enhances disparity estimation accuracy. 

Teacher-student methods have been also considered. Shi et al.\cite{shiBidirectionalSemiSupervisedDualBranch2023} propose a dual-branch CNN-based teacher-student model, designed to further improve disparity estimation accuracy in surgical settings. This framework jointly trains the teacher-student network and a confidence network in a bidirectional semi-supervised manner, where each branch predicts disparity probability distributions, disparity values, and confidence maps. While this approach achieves high accuracy and robustness, it significantly increases computational demands. 

\section{METHODOLOGY}

\subsection{Network Architecture Overview}

\begin{figure*}[!t]
    \centering
    \includegraphics[width=0.6\linewidth]{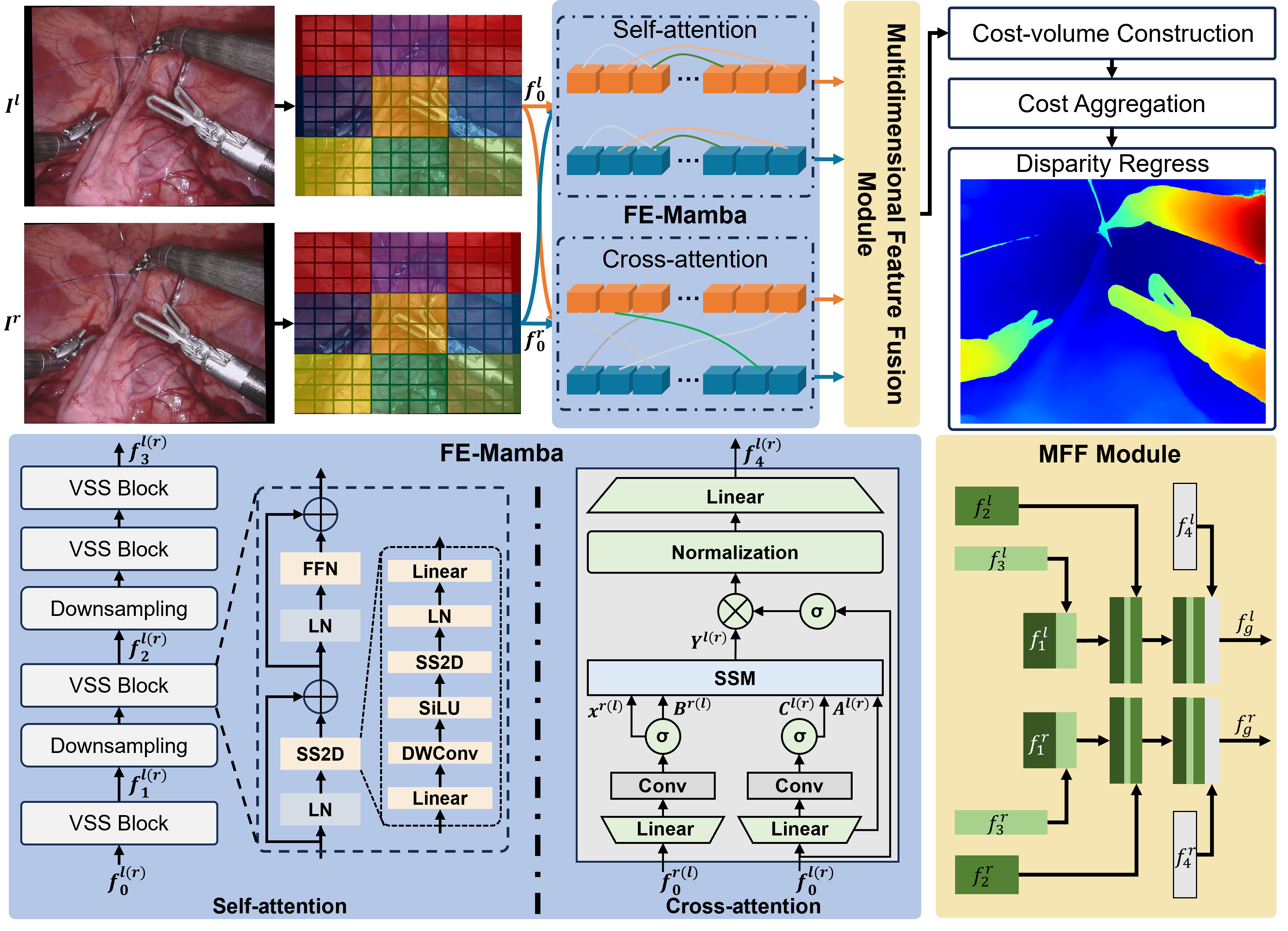}
    \caption{The overall architecture of StereoMamba, which consists of the FE-Mamba module for feature extraction, the cost volume construction via the MFF module, and the cost volume aggregation for disparity estimation.}
    \label{fig:overall}
\end{figure*}

The overall architecture of our StereoMamba is illustrated in Fig.~\ref{fig:overall}. StereoMamba first applies a 2D convolution with a kernel size of 4 and a stride of 4 to downsample the rectified left and right images $I^{l(r)} \in \mathbb{R}^{H\times W\times 3}$, resulting in corresponding feature maps $f_0^{l(r)} \mathbb{R}^{\in \frac{H}{4} \times \frac{W}{4} \times C_0}$, where the superscript of $*^l$ and $*^r$ represent the left and right image or feature map. $H$, $W$ and $C_i$ denote the height, width and number of channels. To leverage global context information within and between the stereo image pair, we propose the FE-Mamba module to perform self- and cross-attention. Accordingly, the MFF module is proposed to effectively combine the self- and cross-attention features. We then follow \cite{guoGroupwiseCorrelationStereo2019} and split features into groups along the channel dimension to compute correlation maps for each group and construct a cost volume. Finally, a cost volume aggregation network is used to regress the disparity value.

\subsection{Feature Extraction Mamba}
The FE-Mamba module contains the components of self-attention and cross-attention. For self-attention, we leverage the four-way scanning strategy of VMamba\cite{liuVMambaVisualState2024} to process individual images. To model cross-attention between image pairs, we design a new visual component inspired by the Mamba2 architecture\cite{daoTransformersAreSSMs2024}.

\subsubsection{Self-attention}
VMamba consists of four stacks of Visual State-Space (VSS) blocks and downsampling layers that process multi-scale features. It utilizes 2D Selective Scan (SS2D) layers, depthwise convolutions (DWConv), SiLU activation, and feed-forward networks (FFN) to capture global features within image efficiently. These operations are interleaved with layer normalization (LN) and linear transformations, ensuring stable training and improved feature representations. Unlike the original VMamba, we eliminate the downsampling block from the final stack, since the low resolution feature output does not facilitate the subsequent upsampling process and leads to reduced stereo matching accuracy. The output features are denoted as $f_1^{l(r)}\in \mathbb{R}^{\frac{H}{4} \times \frac{W}{4} \times C_1}$, $f_2^{l(r)}\in \mathbb{R}^{\frac{H}{8} \times \frac{W}{8} \times C_2}$, and $f_3^{l(r)}\in \mathbb{R}^{\frac{H}{16} \times \frac{W}{16} \times C_3}$.

\subsubsection{Cross-attention}
In Transformers, by far the most commonly used attention mechanism is softmax attention, defined as: $Y=softmax(QK^T)\cdot V$.
A linear attention mechanism can be utilized to remove the softmax by folding it into a kernel feature map\cite{katharopoulos2020transformers}, and using the associativity of matrix multiplication to rewrite $(QK^T)\cdot V =Q\cdot (K^T V )$.
In this case, a mask can be incorporated into the left-hand side as $(L\circ QK^T)\cdot V$, where $L$ is the lower-triangular 1’s matrix, and $\circ$ denotes element-wise product \cite{katharopoulos2020transformers}. The final attention calculation can be written as:
\begin{equation}
\label{equ:transformer}
Y=(L\circ QK^T)\cdot V.
\end{equation}
Considering the preliminaries of SSM that maps a 1-dimensional function or sequence $x(t)\in \mathbb{R} \mapsto y(t)\in \mathbb{R}$ through an implicit hidden state $h(t)\in \mathbb{R}^N$:
\begin{equation}
\begin{aligned}
 h_{t+1} = A_t h_t + B_t  x_t;\space y_t = C_t  h_t,
\end{aligned}
\end{equation}
by definition $h_0 = B_0x_0$,:
\begin{equation}
\begin{aligned}
h_{t}&=A_t\cdots A_1 B_0 x_0+\cdots +A_tB_{t-1}x_{t-1}+B_tx_t \\
 &= \sum_{s=0}^{t}A_{t:s}^{\times}B_sx_s,\\
\end{aligned}
\end{equation}
where $\space A_{t:s}^{\times} =A_tA_{t-1}\cdots A_{s+1},\space A_{s:s}^{\times} = I$. Then the output can be written as:
\begin{equation}
y_{t} = \sum_{s=0}^{t}C_{t}^{T}A_{t:s}^{\times}B_sx_s.
\end{equation}
Next we define a lower-triangular matrix $M_{t,s} = C_{t}^{T}A_{t:s}^{\times}B_s$, where $t$ and $s$ denote the row and column index of the matrix: 
\begin{equation}
M=\begin{bmatrix}
C_0^TB_0 & 0 & 0 & 0 & \cdots \\
C_1^TA_{1:0}^{\times}B_0 & C_1^TB_1 & 0 & 0 & \cdots \\
C_2^TA_{2:0}^{\times}B_0 & C_2^TA_{2:1}^{\times}B_1 & C_2^TB_2 & 0& \cdots\\
\vdots & \vdots & \vdots & \ddots \\
C_t^TA_{t:0}^{\times}B_0 & C_t^TA_{t:1}^{\times}B_1 & C_t^TA_{t:2}^{\times}B_2 & \cdots & C_t^TB_t^T
\end{bmatrix},
\end{equation}
then the output can be rewritten in the matrix format:
\begin{equation}
Y = Mx.
\end{equation}
Each element $M_{t,s}$ can be factorized as:
\begin{equation}
M_{t,s} = C_t^T A_{t:s}^{\times} B_s = (CB^T)_{t,s} \cdot A_{t:s}^{\times},
\end{equation}
where $(CB^T)_{t,s} := C_t^T B_s$. This suggests that $M$ can be expressed as the element-wise product between two matrices:
a matrix $CB^T$ whose $(t,s)$-th element is $C_t^T B_s$, and a lower-triangular matrix $L$ whose $(t,s)$-th element is $A_{t:s}^{\times}$. Specifically, we define:
\begin{equation}
\begin{aligned}
L_{t,s}=\left\{\begin{matrix}
A_{t:s}^{\times}, &  t\geqslant s \\
0, &  t<s \\
\end{matrix}\right.
\end{aligned}
\end{equation}
Thus, we obtain a concise expression for the output:
\begin{equation}
\label{equ:ssm}
Y = (L\circ CB^T)\cdot x.
\end{equation}
This equation is formally identical to Eq.~\ref{equ:transformer}, thus establishing a mathematically unified formulation of Mamba and Transformer architectures. For the sake of brevity, we write this form of Eq.~\ref{equ:ssm} below as $Y = SSM(A, B,C,x)$.

Based on this, we design the cross-attention component of the FE-Mamba module, as shown in Fig.~\ref{fig:overall} (middle bottom). Given input feature maps $f_0^{l(r)}$, linear transformations are first applied to adjust their dimensions. Transformed features are then processed through convolutional layers, parameterized by $x^{r(l)}$, $B^{r(l)}$ and $C^{l(r)}$.  The resulting features are then passed into the SSM, which captures long-range spatial dependencies between stereo images. The SSM generates intermediate outputs $Y^{l(r)}$:
\begin{equation}
\begin{aligned}
Y^{l(r)} = SSM(A^{l(r)}, B^{r(l)}, C^{l(r)}, x^{r(l)}).
\end{aligned}
\end{equation}
The final refined features $f_4^{l(r)}$ are obtained by applying a GeLU activation $\sigma$ to $f_0^{l(r)}$, followed by a root mean squared normalization and a linear transformation.
\begin{equation}
\begin{aligned}
f_4^{l(r)} &= Linear\{RMSNorm\{Y^{l(r)}, \sigma(f_0^{l(r)})\}\}.
\end{aligned}
\end{equation}

\subsection{Cost Volume Construction with Multidimensional Feature Fusion}

To effectively fuse self-attention features ($f_1^{l(r)}$, $f_2^{l(r)}$, $f_3^{l(r)}$) and cross-attention features ($f_4^{l(r)}$), we propose the MFF module. The feature map $f_3^{l(r)}$ from the final VSS block of the self-attention branch is up-sampled using a transposed convolutional layer with ReLU activation to match the dimensions of $f_1^{l(r)}$, and then concatenated with $f_1^{l(r)}$. The concatenated features are then passed through a transposed convolution and concatenated with $f_2^{l(r)}$, followed by the same convolution and activation. Finally, the merged self-attention features are concatenated with the cross-attention features $f_4^{l(r)}$ to obtain the multidimensional feature $f_g^{l(r)}\in \mathbb{R}^{\frac{H}{4},\frac{W}{4},{C_g}}$.

We then divide the multidimensional feature $f_g^{l(r)}$ along the channel dimension into groups and compute correlation maps for each group. All channels are evenly divided into $N_{g}$ groups, therefore each feature group has $C_{g}/N_{g}$ channels. The group-wise correlation is computed as:
\begin{equation}
\begin{aligned}
 C_{gwc}(d,x,y,i) = \frac{1}{C_{g}/N_{g}}\left \langle f_g^{l,i}(x,y), f_g^{r,i}(x,y) \right \rangle   
\end{aligned}
\end{equation}
where $\left \langle \cdot , \cdot\right \rangle $ denotes the inner product, and $f_g^{l,i} $, $f_g^{r,i} $ represent the \textit{i}-th feature group of the left and right multidimensional features, respectively. 

\subsection{Loss Function}
Four outputs are obtained in the cost aggregation. For each output, two 3D convolutions are employed to produce a 1-channel volume, which we then up-sample and convert into a probability volume using softmax along the disparity dimension. For each pixel, we have a $D_{max}$-length vector which contains the probability $p$ for all disparity levels. Finally, the predicted disparity value is computed by the soft-argmin function,
 \begin{equation}
\begin{aligned}
 \hat{d} =\sum_{k=0}^{D_{max}-1 }k\cdot p_{k}  ,
\end{aligned}
\end{equation}
where $k$ and $p_k$ denote the candidate disparity and corresponding probability. The four predicted disparity maps ($\hat{d_{0} }$, $\hat{d_{1} }$, $\hat{d_{2} }$, and $\hat{d_{3} }$) are used to formulate the overall loss as:
 \begin{equation}
\begin{aligned}
Loss=\sum_{i=0}^{3}w_{i}\cdot L_{smooth_{L1}} (\hat{d_{i} }, d),
\end{aligned}
\end{equation}

where $w_{i}$ denotes the coefficients for the $i$th disparity prediction and $d$ denotes the ground-truth disparity map. The smooth L1 loss $ L_{smooth_{L1}}$ is defined as:
\begin{equation}
\begin{aligned}
 L_{smooth_{L1}} (x,y)=\begin{cases}0.5(x-y)^{2}   & \text{ if } \space\left | x-y \right | <1 \\\left | x-y \right | -0.5  & \text{ if }  \space otherwise\end{cases}
\end{aligned}
\end{equation}

\section{MODEL DEVELOPMENT}
\subsection{Datasets and Evaluation Metrics}
\textbf{SceneFlow}\cite{mayerLargeDatasetTrain2016} is a synthetic stereo collection that includes three subsets: Flyingthings3D, Driving, and Monkaa. The resolution is 960×540 and dense disparity maps are provided as ground truth. Following the original splitting strategy, we use the cleanpass category, selecting $35,454$ image pairs for training and $4,370$ for testing.

\textbf{SCARED} is a surgical video dataset, from the MICCAI 2019 Endovis challenge, featuring depth maps of porcine abdominal anatomy captured using a da Vinci Xi endoscope and structured light projectors\cite{allanStereoCorrespondenceReconstruction2021}. It includes 7 training subsets and 2 testing subsets, each containing 4 videos and one image pair at 1280×1024 resolution. Due to calibration errors in subsets 4 and 5 and synchronization issues between RGB videos and depth maps, we use only the first keyframes from subsets 1, 2, 3, 6, and 7, yielding 25 training image pairs. Evaluation is conducted on 2 test subsets, each with 5 keyframes (4 video sequences and 1 image pair), totalling 5909 image pairs.

\textbf{Robotics Instrument Segmentation (RIS\_2017) Dataset}\cite{allan20192017roboticinstrumentsegmentation} is released as part of the MICCAI 2017 Robotic Instrument Segmentation Challenge. This dataset is generated from 10 abdominal porcine operations recorded using the da Vinci Xi system and provides rectified stereo frame data with a resolution of 1280 × 1024. The dataset is sampled at a rate of 1 Hz, yielding 3,000 stereo image pairs.

\textbf{StereoMIS}\cite{hayoz2023pose} is a dataset used for Simultaneous Localization and Mapping (SLAM) in endoscopic surgery, without any disparity or depth ground truth. This dataset is collected using the da Vinci Xi surgical robot. It consists of 10 stereo video sequences recorded at 1280 × 1024 resolution. We extract one frame per ten frames, yielding 12,180 stereo image pairs.

Both RIS\_2017 and StereoMIS have no disparity ground truth. They are used to evaluate zero-shot generalization.

\textbf{Evaluation Metrics.} Following\cite{psychogyiosMSDESISMultitaskStereo2022, guoGroupwiseCorrelationStereo2019}, we use established disparity evaluation metrics: End-Point Error (EPE), Bad2, Bad3, Bad5, depth MAE, and inference speed. EPE represents the mean absolute error between the ground truth and predicted disparity values. Bad2, Bad3 and Bad5 indicates the percentage of pixels where the estimated disparity deviates by more than 2 pixels, 3 pixels, 5 pixels from the ground truth. Depth MAE, measured in millimeters, quantifies the mean absolute error in depth estimation. Inference speed is measured in frames per second (FPS). In all performance metrics, lower values indicate better results. To evaluate the generalization ability on the two datasets without ground truth, we adopt\cite{yan2020disparity} and warp left images using disparity maps to synthesize right images, then compare them with actual right images. The similarity is assessed using three widely used metrics: Structural Similarity Index (SSIM), Peak Signal-to-Noise Ratio (PSNR), and Learned Perceptual Image Patch Similarity (LPIPS).

\begin{table*}[!t]
\caption{Evaluation on SCARED Test subsets. $Kn$ represents the n\textit{th} keyframe of each subset.The table reports depth MAE for each keyframe, along with the mean depth MAE, Bad2, Bad3, Bad5, and EPE across all test keyframes. \textbf{Bold}: Best, \underline{Underline}: Second-best.}
\label{tab:leaderboard}
\centering
\resizebox{2\columnwidth}{!}{
\begin{tabular}{|c|c|c|c|c|c|c|c|c|c|c|c|c|c|c|c|c|}
\hline
\multirow{2}{*}{Methods} & \multicolumn{5}{c|}{Test subset 1 (Depth MAE: mm $\downarrow$)} & \multicolumn{5}{c|}{Test subset 2 (Depth MAE: mm $\downarrow$)} & \multirow{2}{*}{\makecell[c]{mean Depth \\MAE (mm)$\downarrow$}}&  \multirow{2}{*}{Bad2 (\%)$\downarrow$}&\multirow{2}{*}{Bad3 (\%)$\downarrow$}  &\multirow{2}{*}{Bad5 (\%)$\downarrow$}& \multirow{2}{*}{EPE (px)$\downarrow$}&\multirow{2}{*}{\makecell[c]{Inference \\speed (FPS)$\uparrow$}}\\ 
\cline{2-11}
& K0& K1& K2& K3& K4& K0& K1& K2& K3& K4&  &   &&& &\\ \hline
GwcNet-gc\cite{guoGroupwiseCorrelationStereo2019}& 8.60& 2.63& \underline{1.49}& \underline{2.06}& 0.66& 4.25& 1.00& 3.41& \textbf{1.44}& \underline{0.35}&  2.59&  41.55&27.28 &14.07& 2.67&3.23\\ \hline
ACVNet\cite{xuAttentionConcatenationVolume2022}& 8.56& 2.59& \underline{1.49}& 2.08& 0.68& 4.23& 0.96& 3.46& \textbf{1.44}& \textbf{0.33} & 2.58&  \textbf{40.66}&\textbf{26.50} &\underline{13.85}& 2.70 &2.70\\ \hline
RAFT-Stereo\cite{lipsonRAFTStereoMultilevelRecurrent2021} & 8.45& 2.51& 1.75& \textbf{2.03}& 0.73& \textbf{4.14}& \textbf{0.89}& \textbf{3.06}& 1.52& 0.68&  2.58&  44.91&29.80 &14.19& 2.83&5.00\\ \hline
IGEV-Stereo\cite{xuIterativeGeometryEncoding2023}& \underline{8.27}& \underline{2.35}& 1.73& 2.18& \textbf{0.56}& 4.41& 0.97& 3.22& 1.54& 0.37&  \underline{2.56}&  42.15&27.28 & \textbf{13.82}& \underline{2.65}&2.44\\ \hline
Selective-Stereo\cite{wang2024selective}& 8.44& 2.44& 1.62& 2.27& 0.70& 4.30& 0.94& 3.34& 1.78& 0.66& 2.65&  43.48&28.36 &14.17& 2.81&1.96\\ \hline
MSDESIS\cite{psychogyiosMSDESISMultitaskStereo2022}& 8.42& 2.59& 2.05& 3.02& 1.01& 4.68& 1.18& 3.34& 1.61& 0.46&  2.84&  44.71&29.81 &16.14& 3.06&\textbf{100.00}\\
\hline
Shi et al.\cite{shiBidirectionalSemiSupervisedDualBranch2023}& \textbf{7.61}& \textbf{2.10}& 1.97& 2.66& \underline{0.65}& 4.75& 1.18& 2.96& 1.71& 0.36  & 2.60 &  44.41&28.54  & 14.28 & 2.74&1.08\\ 
\hline
StereoMamba (Ours) & 8.57& 2.61& \textbf{1.46}& \textbf{2.03}& 0.77& \underline{4.19}& \underline{0.92}& \underline{3.07}& \underline{1.46}& 0.41& \textbf{2.55}&  \underline{41.49}&\underline{26.99} &13.88& \textbf{2.64}&\underline{21.28}\\ \hline
\end{tabular}
}
\end{table*}

\begin{figure*}[!ht]
    \centering
    \includegraphics[width=\linewidth]{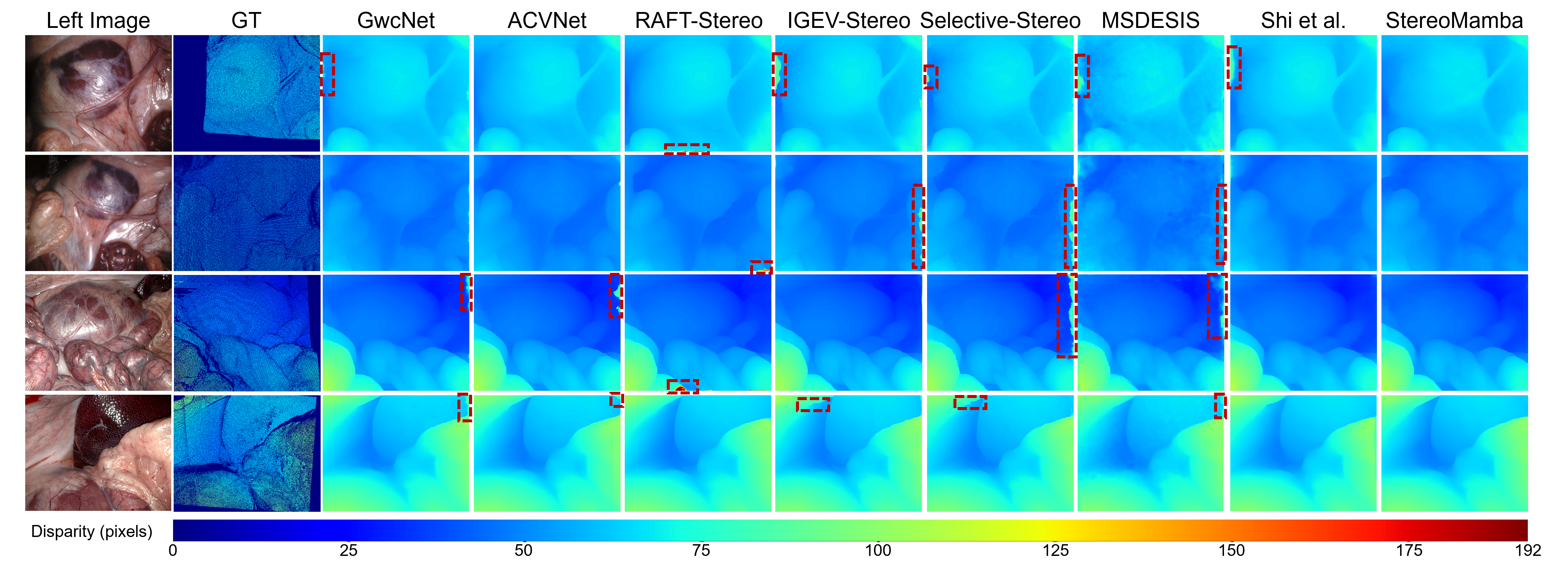}
    \caption{Qualitative results on SCARED. The first column indicates the rectified left image, the second column indicates the ground truth disparity, and the other columns are estimated disparity maps from GwcNet\cite{guoGroupwiseCorrelationStereo2019}, ACVNet\cite{xuAttentionConcatenationVolume2022}, RAFT-Stereo\cite{lipsonRAFTStereoMultilevelRecurrent2021}, IGEV-Stereo\cite{xuIterativeGeometryEncoding2023}, Selective-Stereo\cite{wang2024selective}, MSDESIS\cite{psychogyiosMSDESISMultitaskStereo2022}, Shi et al.\cite{shiBidirectionalSemiSupervisedDualBranch2023} and our StereoMamba. All methods share the same disparity colorbar, ranging from 0 to 192 pixels. Additional visualization results are provided in the supplementary material.}
    \label{fig:scared}
\end{figure*}

\subsection{Implementation Details}
All experiments are implemented in PyTorch on a single Nvidia RTX A6000 GPU with 48 GB of memory. Initially, StereoMamba is pre-trained on the SceneFlow (cleanpass) dataset for 40 epochs with a batch size of 14. We adopt the AdamW optimizer with $\beta_1$ = 0.9, $\beta_2$ = 0.999, and a weight decay of 1e-4. We employ a one-cycle learning rate policy with a maximum learning rate of 2e-4, keeping the other hyperparameters at PyTorch’s default values. Data augmentation includes random crops of size 512 × 256 and color normalization based on each dataset's statistics. The maximum disparity value is $D_{max}$ = 192. The coefficients of four outputs are set as $w_0$ = 0.5, $w_1$ = 0.5, $w_2$ = 0.7, $w_3$ = 1.0 following\cite{guoGroupwiseCorrelationStereo2019}. Subsequently, the model is fine-tuned on the SCARED dataset with a constant learning rate of 1e-3 for 150 epochs.

\section{EXPERIMENTS and RESULTS}

\subsection{Comparison with SOTA on the SCARED Benchmark}

Following the recommended evaluation protocol from\cite{allanStereoCorrespondenceReconstruction2021}, the evaluation is performed on all frames except those in which more than 90\% of the ground truth disparity maps are empty. Seven SOTA methods including, the baseline method GwcNet\cite{guoGroupwiseCorrelationStereo2019}, one cost volume-optimization method ACVNet\cite{xuAttentionConcatenationVolume2022}, three iterative-optimization methods RAFT-Stereo\cite{lipsonRAFTStereoMultilevelRecurrent2021}, IGEV-Stereo\cite{xuIterativeGeometryEncoding2023} and Selective-Stereo\cite{wang2024selective}, one multi-tasking method MSDESIS\cite{psychogyiosMSDESISMultitaskStereo2022} and one semi-supervised teacher-student method Shi et al.\cite{shiBidirectionalSemiSupervisedDualBranch2023}, are chosen for comparison. For methods that follow the standard pre-training on SceneFlow and fine-tuning pipeline\cite{guoGroupwiseCorrelationStereo2019, xuAttentionConcatenationVolume2022, lipsonRAFTStereoMultilevelRecurrent2021,xuIterativeGeometryEncoding2023, wang2024selective, psychogyiosMSDESISMultitaskStereo2022}, we fine-tune them on the same SCARED dataset as ours to ensure a fair comparison. Since Shi et al.\cite{shiBidirectionalSemiSupervisedDualBranch2023} only provides inference code and weights trained on SCARED, we use the released model to directly generate disparity maps.

Table~\ref{tab:leaderboard} lists comparative results of StereoMamba against the SOTA approaches. Our StereoMamba outperforms all competing methods on K2 and K3 videos of Test Subset 1, achieving depth MAEs of 1.46 mm and 2.03 mm, respectively. For Test Subset 2, StereoMamba ranks second-best on K0, K1, K2, and K3 videos, with performance differences ranging from just 0.01 mm to 0.05 mm compared to the top methods\cite{guoGroupwiseCorrelationStereo2019,lipsonRAFTStereoMultilevelRecurrent2021,xuAttentionConcatenationVolume2022}. The results demonstrate that StereoMamba achieves comparable or even superior disparity estimation accuracy compared to other SOTA methods.

In summary, StereoMamba outperforms all competing methods on the entire SCARED dataset in terms of both mean depth MAE (2.55 mm) and EPE (2.64 px). It also ranks as second-best on Bad2 (41.49\%) and Bad3 (26.99\%), trailing the top-performing ACVNet\cite{xuAttentionConcatenationVolume2022} by only 0.83\% and 0.44\%, respectively. For Bad5, StereoMamba reports a value of 13.88\%, just 0.06\% behind the best-performing method IGEV-Stereo\cite{xuIterativeGeometryEncoding2023}. Although other methods perform well on individual keyframes, their overall performance on the entire dataset is inferior to StereoMamba. This consistently strong performance across multiple evaluation metrics highlights StereoMamba’s robustness in handling the challenging areas in SCARED, including specular reflections and textureless regions that are particularly difficult for stereo matching.

Fig.~\ref{fig:scared} shows qualitative results in challenging areas. Notably, compared to the other methods, StereoMamba demonstrates superior performance at image edges and in dark regions, as highlighted by the red dashed-lined boxes. StereoMamba maintains low depth MAE, Bad2, Bad3, Bad5 and EPE, indicating its reliability.

\begin{table*}[!t]
\caption{The evaluation of generalization ability on StereoMIS and RIS\_2017 datasets.}
\label{tab:generalization}
\centering
\resizebox{1.6\columnwidth}{!}{
\begin{tabular}{|c|ccc|ccc|ccc|}
\hline
\multirow{2}{*}{} & \multicolumn{3}{c|}{StereoMIS} & \multicolumn{3}{c|}{RIS\_2017} & \multicolumn{3}{c|}{Average}\\  \hline
& SSIM $\uparrow$ & PSNR $\uparrow$ & LPIPS $\downarrow$ & SSIM $\uparrow$ & PSNR $\uparrow$ & LPIPS $\downarrow$  & SSIM $\uparrow$ & PSNR $\uparrow$ &LPIPS $\downarrow$  \\ \hline
GwcNet-gc\cite{guoGroupwiseCorrelationStereo2019}& 0.9120& 17.1222& 0.2834& 0.8697& 14.5847&0.3539  & 0.8908& 15.8534&0.3187\\ \hline
ACVNet\cite{xuAttentionConcatenationVolume2022}& \underline{0.9141}& \textbf{17.3140}& \textbf{0.2770}& 0.8696& 14.4944&0.3428  & 0.8919& 15.9042&0.3099\\ \hline
RAFT-Stereo\cite{lipsonRAFTStereoMultilevelRecurrent2021}&  0.9108&  17.1873&  \underline{0.2781}&  0.8637&  14.5558&  \textbf{0.3303} & 0.8873& 15.8715&\textbf{0.3042}\\  \hline
IGEV-Stereo\cite{xuIterativeGeometryEncoding2023}& 0.9096& 16.9007& 0.2823& 0.8641& 14.3403&0.3440  & 0.8869& 15.6205&0.3131\\  \hline
Selective-Stereo\cite{wang2024selective} &  0.9110&  16.8081&  0.2888&  \textbf{0.8793}&  \underline{14.7840}&  0.3592  & \underline{0.8952}& 15.7961&0.3240\\  \hline
MSDESIS\cite{psychogyiosMSDESISMultitaskStereo2022}& 0.8842& 15.6194& 0.3263& 0.8526& 14.1028&0.3699  & 0.8684& 14.8611&0.3481\\ \hline
Shi et al.\cite{shiBidirectionalSemiSupervisedDualBranch2023}& 0.9134& 17.2028& 0.2818&  0.8755&  14.6991&  \underline{0.3353} & 0.8945& \underline{15.9509}&\underline{0.3086}\\ \hline
StereoMamba (Ours) & \textbf{0.9149}& \underline{17.3054}& 0.2786&  \underline{0.8790}&  \textbf{14.8468}&  0.3431  & \textbf{0.8970}& \textbf{16.0761}& 0.3109\\  \hline
\end{tabular}
}
\end{table*}

\begin{figure*}[!h]
    \centering
    \includegraphics[width=0.9\linewidth]{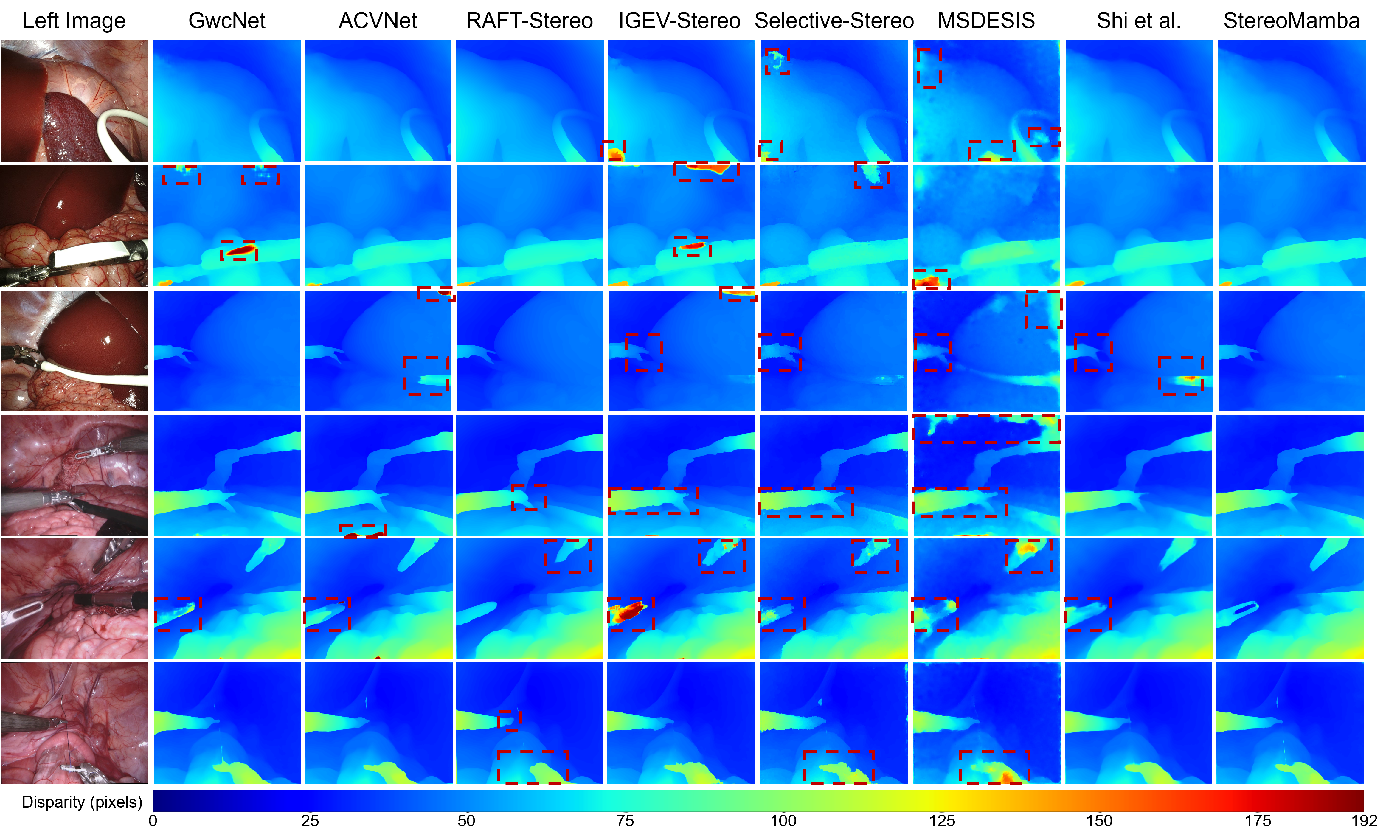}
    \caption{Qualitative results on StereoMIS (first 3 rows) and RIS\_2017 (last 3 rows). The first column indicates the rectified left image, the other columns are estimated disparity maps from GwcNet\cite{guoGroupwiseCorrelationStereo2019}, ACVNet\cite{xuAttentionConcatenationVolume2022}, RAFT-Stereo\cite{lipsonRAFTStereoMultilevelRecurrent2021}, IGEV-Stereo\cite{xuIterativeGeometryEncoding2023}, Selective-Stereo\cite{wang2024selective}, MSDESIS\cite{psychogyiosMSDESISMultitaskStereo2022}, Shi et al.\cite{shiBidirectionalSemiSupervisedDualBranch2023} and our StereoMamba. All methods share the same disparity colorbar, ranging from 0 to 192 pixels. Additional visualization results are provided in supplementary material.}
    \label{fig:stereomis_and_ris}
\end{figure*}

More importantly, StereoMamba delivers real-time performance with an inference speed at 21.28 FPS, significantly outperforming cost volume methods\cite{guoGroupwiseCorrelationStereo2019, xuAttentionConcatenationVolume2022}, iterative-optimization methods\cite{lipsonRAFTStereoMultilevelRecurrent2021, xuIterativeGeometryEncoding2023, wang2024selective}, and the semi-supervised teacher-student method\cite{shiBidirectionalSemiSupervisedDualBranch2023}, which operate only at 1.08 to 5.00 FPS. Although MSDESIS\cite{psychogyiosMSDESISMultitaskStereo2022} achieves real-time inference, it suffers from poor disparity estimation accuracy. Overall, StereoMamba exhibits strong robustness while effectively balancing disparity estimation accuracy and inference speed, achieving the required trade-off for real-world RAMIS applications.

\subsection{Zero-shot Generalization Results}

Considering the high variability in surgical scenes due to different patient anatomy, hardware utilized and the various surgical applications of RAMIS (e.g. urology, gynaecology), the generalization ability of stereo disparity estimation models is a key performance indicator. We thus evaluate the zero-shot generalization performance of StereoMamba and other methods in unseen real-world, \textit{in-vivo} surgical scenes. Following a zero-shot setting, all methods are trained on SceneFlow, fine-tuned on SCARED, and directly tested on the RIS\_2017 and StereoMIS datasets.

As shown in Table~\ref{tab:generalization}, our approach achieves comparable or superior performance against SOTA methods across both \textit{in-vivo} datasets. On StereoMIS, StereoMamba attains the best SSIM score (0.9149), the second-best PSNR score (17.3054), and an LPIPS score of 0.2786, just 0.0016 higher than the best result. In RIS\_2017, StereoMamba achieves the best PSNR score (14.8468), the second-best SSIM score (0.8790), and an LPIPS score of 0.3431, only 0.0128 higher than the top method. On average, StereoMamba achieves the best SSIM (0.8970) and PSNR (16.0761), with LPIPS being\cite{lipsonRAFTStereoMultilevelRecurrent2021} only 0.0067 higher than the best method. Evidently, StereoMamba demonstrates strong generalization capability in generating reliable disparity maps on unseen \textit{in-vivo} datasets.

Example results are presented in Fig.~\ref{fig:stereomis_and_ris}, with notable disparity estimation errors highlighted in dashed red boxes. GwcNet\cite{guoGroupwiseCorrelationStereo2019} struggles with dark regions and specular reflections from instruments. ACVNet\cite{xuAttentionConcatenationVolume2022} improves performance in dark areas but still suffers from specular reflections. RAFT-Stereo\cite{lipsonRAFTStereoMultilevelRecurrent2021} handles these challenges better but loses some instrument details on the RIS\_2017 dataset. IGEV-Stereo\cite{xuIterativeGeometryEncoding2023} and Selective-Stereo\cite{wang2024selective} struggle with image edges, dark regions, and specular reflections, leading to unsmooth instrument boundaries. MSDESIS\cite{psychogyiosMSDESISMultitaskStereo2022} performs the worst, producing disparity maps with significant noise, while Shi et al.\cite{shiBidirectionalSemiSupervisedDualBranch2023} shows slight improvement but remains affected by specular reflections. In contrast, StereoMamba demonstrates strong robustness against these challenges, highlighting its superiority for zero-shot generalization compared to other methods.


\subsection{Ablation Study}
To verify the effectiveness of our proposed modules, we take GwcNet\cite{guoGroupwiseCorrelationStereo2019} as the baseline and replace its ResNet backbone with our FE-Mamba and MFF modules. As shown in Table~\ref{tab:ablation_study}, replacing ResNet with FE-Mamba alone (StereoMamba-base) leads to notable improvements: EPE is reduced by 0.1651 px, Bad2 by 0.8\%, Bad3 by 1.04\%, and Bad5 by 0.45\%, while inference speed significantly increases from 4.76 FPS to 27.78 FPS, demonstrating the efficiency and effectiveness of FE-Mamba. With the addition of the MFF module, the full StereoMamba model achieves further improvements, reducing Bad2 by an additional 0.09\%, Bad3 by 0.01\%, and Bad5 by 0.04\%. Although the inference speed slightly decreases to 27.03 FPS, the model runs at 21.28 FPS on the 1280 $\times$ 1024 image pairs of SCARED, the trade-off between inference speed and accuracy is acceptable. The inference speed of 21.28 FPS is sufficient for real-time depth estimation during surgery, and the marginal improvement in speed beyond this threshold offers limited practical benefit. From this perspective, we believe that the modest reduction in FPS is fully justified by the improved accuracy.

\begin{table}[!t]
\caption{Ablation study results of proposed networks on the Cleanpass of SceneFlow dataset. All metrics are for 960×540 inputs on a single Nvidia RTX 4090 GPU.}
\centering
\resizebox{\columnwidth}{!}{
\begin{tabular}{|c|c|c|c|c|c|c|c|c|}    \hline
Method  &ResNet & FE-Mamba& MFF & EPE (px) $\downarrow$ & Bad2 (\%) $\downarrow$&Bad3 (\%)$\downarrow$&Bad5 (\%)$\downarrow$& \makecell[c]{Inference \\ speed (FPS)$\uparrow$}\\ \hline
GwcNet\cite{guoGroupwiseCorrelationStereo2019} &$\surd$ &  &  &  0.7691  & 4.37&3.30 &2.25&  4.76\\ \hline
StereoMamba-base&& $\surd$ &  &   \textbf{0.6044}   & 3.57&2.26 &1.80&  \textbf{27.78} \\ \hline
StereoMamba&& $\surd$ & $\surd$ & \textbf{0.6044}   & \textbf{3.48}&\textbf{2.25} &\textbf{1.76}&  27.03\\ \hline

\end{tabular}}
\label{tab:ablation_study}
\end{table}

\section{CONCLUSIONS}

In this paper, we propose StereoMamba, the first method to explore SSM for disparity estimation in RAMIS. We design a specialized FE-Mamba module to perform both self-attention and cross-attention within and cross stereo images, effectively encoding long-range spatial features, which are then seamlessly integrated using our novel MFF module. The fused multidimensional features are processed by a group-wise correlation-based decoder to generate the final disparity map. On the SCARED benchmark, StereoMamba achieves SOTA performance with an EPE of 2.64 px and a Depth MAE of 2.55 mm. It also delivers competitive results on Bad2 (41.49\%), Bad3 (26.99\%) and Bad5 (13.88\%), while maintaining a real-time inference speed of 21.28 FPS for 1280 × 1024 image pairs. Compared to existing methods, it produces smooth and stable disparity estimations, even in challenging regions such as specular reflections and textureless areas. This balance between accuracy, robustness and inference speed makes StereoMamba well-suited for real-world deployment. 
Additionally, StereoMamba demonstrates strong zero-shot generalization on two unseen \textit{in-vivo} datasets (RIS\_2017, StereoMIS), achieving an SSIM of 0.8970, PSNR of 16.0761, and LPIPS of 0.3109 when comparing synthesized right images with the actual ones.





\bibliographystyle{IEEEtran}
\bibliography{Mylibrary.bib}

\end{document}